\documentclass{article}

\usepackage{arxiv}

\usepackage[utf8]{inputenc} % allow utf-8 input
\usepackage[T1]{fontenc}    % use 8-bit T1 fonts
\usepackage{hyperref}       % hyperlinks
\usepackage{url}            % simple URL typesetting
\usepackage{booktabs}       % professional-quality tables
\usepackage{amsfonts}       % blackboard math symbols
\usepackage{nicefrac}       % compact symbols for 1/2, etc.
\usepackage{microtype}      % microtypography
\usepackage{lipsum}		% Can be removed after putting your text content
\usepackage{graphicx}
\usepackage{natbib}
\usepackage{doi}

\usepackage{subfigure}
\usepackage{svg}
\usepackage{float}
\usepackage{booktabs} % for professional tables\
\usepackage{algorithm}
\usepackage{algorithmic}

\usepackage{amsmath}
\usepackage{amssymb}
\usepackage{mathtools}
\usepackage{amsthm}
\usepackage[capitalize,noabbrev]{cleveref}

\theoremstyle{plain}

\theoremstyle{definition}

\theoremstyle{remark}

\title{Knowledge Distillation from Neural Networks to Differentiable Logic Networks}

\author{ Hee Yun Suh \\
	Dept. of Electrical and Computer Engineering\\
	Princeton University\\
	Princeton, NJ \\
	\texttt{hs9190@princeton.edu} \\
	\And
	Niraj K. Jha \\
	Dept. of Electrical and Computer Engineering\\
	Princeton University\\
	Princeton, NJ \\
	\texttt{jha@princeton.edu} \\
}

\date{}

\renewcommand{\shorttitle}{Knowledge Distillation from Neural Networks to Differentiable Logic Networks}

\begin{document}
\maketitle

\begin{abstract}
	Differentiable logic networks (DLNs) are interpretable, edge-friendly classifiers that learn both their logic operators and their connections through gradient descent, achieving accuracy competitive with multi-layer perceptrons at orders-of-magnitude lower inference cost. Their accuracy, however, is bounded by the problems of vanishing gradients, scalability, and lack of an ecosystem, leaving a gap relative to conventional neural networks on several datasets. We introduce knowledge distillation for DLNs to narrow this gap. We distill from a trained multi-layer perceptron teacher into a DLN student, integrating the distillation objective into a two-phase training loop without altering the architecture or the inference procedure. The teacher is discarded after training. Hence, the distilled model retains the interpretability, sparsity, and inference cost of a conventionally trained DLN. We investigate this approach through four variants: a joint hyperparameter search that optimizes the distillation weight and temperature alongside the architecture, a grid search over these two hyperparameters with the architecture fixed, and two single-phase variants that employ distillation during only the neuron-function or only the connection phase. On 20 public tabular classification tasks, distillation improves balanced-class accuracy at negligible cost to gate-level operation counts and model size, though the benefit varies by dataset. Performance improvements of up to 15\% are observed, in some cases surpassing the teacher itself.
\end{abstract}

\section{Introduction}
Interpretable machine learning has increasingly turned from explaining trained black-box models to building models that are interpretable by construction \citep{rudin}. Among the latter, networks built from logic gates have recently emerged as a compelling option for tabular classification. Deep differentiable logic gate networks (DDLGNs) \citep{ddlgn} and differentiable logic networks (DLNs) \citep{dln} replace the weighted sums of a conventional network with real Boolean operators, so that a trained model is a logic network whose decisions can be interpreted directly. These networks attain accuracy competitive with multi-layer perceptrons (MLPs) while executing as sparse Boolean computation at orders-of-magnitude lower inference cost \citep{dln} through training with differentiable relaxation, real-valued logic during optimization, and discretization to hard gates for inference.

The same works also document that the efficiency and transparency come at the cost of logic networks being difficult to train and quick to saturate. Because every layer in the logic network is relaxed through Softmax and sigmoid functions, the gradient signal attenuates as it propagates backward through the stack.  This is a vanishing-gradient effect that both prior works identify as a central training obstacle. It limits the depth at which these networks remain trainable \citep{ddlgn, dln}. The practical consequence is an accuracy ceiling reached early.  In the results presented in \citep{dln}, we observe that on several tabular datasets, a DLN trails a conventional neural network in performance and, unlike neural networks, additional capacity does not reliably close the gap.
Neural networks, meanwhile, benefit from a large and mature ecosystem, such as architectural components \citep{he2016deep}, optimizers \citep{adam}, normalization schemes \citep{batchnorm}, and training infrastructure \citep{pytorch}. This ecosystem has made scaling to higher accuracy comparatively routine. The interpretable logic network is thus easier to trust but harder to improve.

We observe that this asymmetry is precisely the setting in which knowledge distillation (KD) is useful. KD was introduced as a means to compress the capability of a large model into a smaller one \citep{kd_hinton}. 
Subsequent work has transferred teacher capability to students that are structurally constrained, such as quantized networks \citep{polino2018model, mishra2017apprentice}, binary networks \citep{leroux2020training}, and decision trees \citep{frosst2017distilling}. A DLN is also a constrained student, as its expressiveness is bounded not by parameter count but by its discrete operators and sparse wiring. Prior work has translated trained networks to logic representations by fitting intermediate models to quantized activations and compiling to AND-Inverter graphs \citep{brudermueller2020making}. However, this is a post-hoc translation of a fixed logic network rather than a distillation objective, and it inherits an accuracy loss during quantization and compilation.  To our knowledge, distillation of neural networks into relaxation-trained logic networks has not been studied, and whether the soft-target signal that benefits conventional students transfers to a model so tightly constrained by discrete operators and sparse wiring is an open question.

In this article, we distill knowledge from a trained neural-network teacher into a DLN student, motivated by DLN's specific training pathology rather than by the need for compression. A one-hot label provides only a sparse error signal at the aggregation layer, where the DLN's backward pass originates. A teacher's softened distribution supplies a denser, better-scaled gradient at that layer. The teacher participates only during training and is discarded at inference; hence, the distilled DLN retains the interpretability, sparsity, and inference cost of a conventionally trained DLN while inheriting accuracy it cannot otherwise reach.

Our main contributions are as follows.

\begin{itemize}
\item We formulate a KD methodology for DLNs, integrating the distillation objective into its two-phase training procedure without altering the architecture or the inference procedure.
\item We compare several distillation variants, including a dedicated grid search and distillation applied, respectively, to the neuron-function phase, the connection phase, or both, to determine which yields the most accurate distilled DLN.
\item We evaluate the methodology on 20 public tabular datasets, reporting accuracy alongside gate-level operation counts and model size, and show that distillation improves accuracy without increasing inference cost.
\end{itemize}

The remainder of this article is organized as follows. Section II discusses related work. Section III describes the distillation protocol, model training, and implementation details. Section IV presents experimental results. Section V provides concluding remarks and directions for future research.

\section{Related Work}

In this section, we review prior work on interpretable and efficient machine learning, DLNs and the relaxation techniques that make them trainable, KD, and distillation of knowledge into structurally constrained student models.

\subsection{Interpretable and Efficient Machine Learning}
A substantial body of work seeks to make model predictions interpretable. A few notable works on ad-hoc methods applicable to already-trained black-box models include LIME \citep{lime}, Shapley additive explanations \citep{shapley}, and saliency maps \citep{saliencymaps}. These methods attribute a prediction to individual input features without constraining the model that produced it. \citep{rudin} argues that such post-hoc explanations are frequently unreliable in exactly the high-stakes settings that motivate them, and advocates instead for models that are inherently interpretable, such as sparse decision trees \citep{gosdt} and scoring systems \citep{slim}. Such models, however, often accept a reduction in accuracy in exchange for interpretability. It is this gap that our work seeks to narrow. A separate line of work targets efficiency rather than interpretability, reducing inference cost through pruning \citep{han2015pruning}, quantization \citep{hubara2018quantized}, binary neural networks \citep{hubara2016binarized, courbariaux2016bnn}, and specialized hardware, such as field-programmable gate arrays \citep{finn}. These techniques compress a black box into a smaller or faster one, but do not make it more interpretable. DLNs are notable in claiming both interpretability by construction and low inference cost.

\subsection{Differentiable Logic Gate Networks}
The immediate predecessor to our work is the deep differentiable logic gate network (DDLGN) \citep{ddlgn}, which trains a network of two-input Boolean logic gates through continuous relaxation. In a DDLGN, the wiring between neurons is randomly initialized and frozen, and each neuron learns a Softmax distribution over the 16 possible two-input Boolean operators. At inference, the distribution is collapsed to its argmax operator, yielding a hard logic network that executes extremely quickly. The DLN \citep{dln} generalizes this construction by making all components of the network differentiable: It learns the connections between neurons in addition to their operators, trains the thresholds that binarize continuous inputs, and accepts continuous features directly rather than through pre-binarized inputs. On tabular datasets, this yields higher accuracy and smaller model size relative to DDLGNs \citep{dln}. Subsequent work has extended logic gate networks along several axes, including convolutional architectures \citep{cdlgn}, recurrent networks \citep{recurrent_lgn}, regression tasks \citep{r_dln}, and time-series classification \citep{tsc_dln}. Our contribution is orthogonal to these architectural extensions, in that we leave the DLN architecture unchanged and instead only modify the training.

The trainability of these networks rests on relaxation techniques imported from several areas. Fuzzy logic and the associated families of T-norms and T-conorms \citep{fuzzy, fuzzylogic} enable Boolean operations to take continuous values in $[0,1]$, so that logic gates become differentiable functions of their inputs. The Softmax function relaxes the discrete selection of an operator, or of an input connection, into a probabilistic mixture over candidates, permitting gradient-based search over what is fundamentally a discrete space. The Straight-Through Estimator \citep{ste} sharpens the forward activations toward their discrete values while enabling gradients to pass through unchanged on the backward pass. These techniques enable training but, as noted, also introduce  stacked-relaxation gradient attenuation that motivates our approach.

\subsection{Knowledge Distillation}
The idea of training a compact model to mimic a larger one dates to \citep{bucilua2006model}, who proposed model compression by fitting a small student to the predictions of a large ensemble. \citep{kd_hinton} formalized this as KD: Rather than training the student on hard labels alone, one softens the teacher's output distribution with a temperature and trains the student to match it. The resulting soft targets convey relative information (also called dark knowledge) about all classes that a one-hot label omits. The temperature-scaled soft-target loss is conventionally weighted by the square of the temperature so that its gradient magnitude remains comparable to that of the hard-label loss. Beyond matching outputs, feature-based methods, such as FitNets \citep{fitnets} and relational distillation \citep{park2019relational}, align the student's intermediate representations with the teacher's, providing additional supervision at hidden layers.  We adopt output-level distillation in this work and note intermediate matching as a direction for extension. A further variant, self-distillation or the born-again setting \citep{furlanello2018born}, uses a teacher of the same architecture as the student and can improve the student's performance beyond that of the teacher. We return to this possibility as an alternative source of teachers in our discussion of future work.

\subsection{Distillation into Constrained Students}
Most relevant to our setting is a line of work that distills not merely into a smaller student but into one whose form is structurally constrained. KD has been used to train quantized and binary neural networks \citep{polino2018model, mishra2017apprentice, leroux2020training}, where the student cannot represent the teacher's continuous weights and must approximate its function under discrete constraints, and to distill neural networks into decision trees \citep{frosst2017distilling}, where the student's hypothesis class is symbolic rather than continuous. A related effort translates trained networks into logic-gate representations for verifiability and interpretability \citep{brudermueller2020making}. In each case, the student's limitation is one of representational form rather than size and distillation serves to fit that form as closely as possible to a more expressive teacher. To our knowledge, distillation into relaxation-trained logic networks has not previously been studied. Unlike the constrained students above, whose limitation is one of precision or a fixed hypothesis class, a DLN is bounded by its discrete operators and sparse two-input wiring, and whether soft-target supervision transfers through that bottleneck is the question we take up in the following section.

%%%%%%%%%%%%% METHODS %%%%%%%%%%%%%%%%%
\section{Methodology}
In this section, we describe our method in detail, beginning with the architecture and training procedure we build upon, followed by the distillation objective and the details of its integration.

\begin{figure*}[ht]
\begin{center}
\centerline{\includegraphics[width=\textwidth]{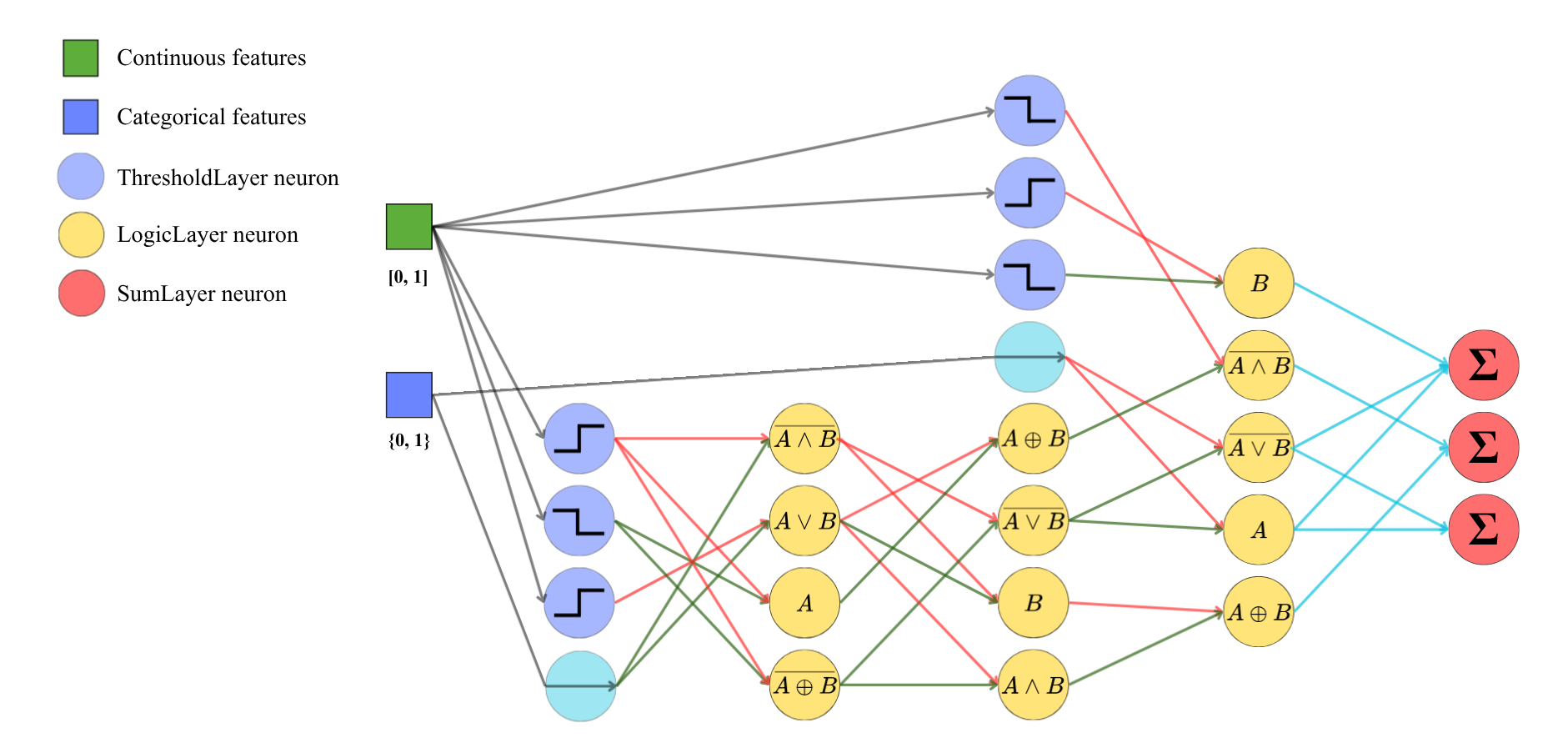}}
\caption{A simple example of our DLN. The model takes binarized continuous variables and categorical variables through a ThresholdLayer. Then, the binary vectors are passed to the LogicLayers. We concatenate the inputs with a ThresholdLayer at every intermediate LogicLayer, i.e. except the first and last one.  At the end, the SumLayer determines the input's class.}
\label{dln_image}
\end{center}
\end{figure*}

We build on the DLN  \citep{dln}, itself an extension of the DDLGN \citep{ddlgn}. As 
Fig.~\ref{dln_image} illustrates, a DLN comprises three components: a ThresholdLayer that binarizes continuous inputs via shifted, scaled sigmoids, a stack of LogicLayers whose neurons apply a Boolean operator to two selected inputs, and a SumLayer that, for each of the $C$ classes, sums the outputs of a subset of last-layer neurons to form a class score. We also implement concatenating inputs, similar to \citep{dln}, where we bring the inputs closer to the output layers by concatenating a ThresholdLayer to every intermediate LogicLayer.  We use subspace search where we search over a set of eight Boolean operations and eight links from the previous layer instead of the 16 logic operations and all links from the previous layer to simplify training and reduce the impact of the vanishing gradient problem. The trainable parameters and per-layer feedforward computations are unchanged from \citep{dln}. Our method only modifies the training objective, leaving the architecture and the inference procedure intact.

Training a DLN requires making its discrete components differentiable, so that both the function of each neuron and the wiring between neurons are learned. Because neuron functions and neuron connections constitute distinct sets of parameters, they are optimized in two alternating phases: Phase I optimizes neuron functions while keeping connections quantized and frozen, and Phase II optimizes the connections while keeping neuron functions quantized and frozen \citep{dln}. After training, both the functions and the connections are discretized to hard binary states for inference.

This layered relaxation is what enables gradient-based training. However, it is also the source of the limitation we address. Every layer passes its signal through a Softmax or sigmoid function. Thus, the resulting attenuation of the backward signal becomes a central obstacle in training accurate logic networks \citep{ddlgn, dln}. A conventional cross-entropy objective compounds the problem: A one-hot label provides a sparse error signal at the SumLayer, which is precisely where the DLN's backward pass originates. Our approach is, therefore, to enrich the signal at that layer rather than alter the network that carries it. We supervise the student with the softened output distribution of a trained teacher network, which conveys relative information about all classes and yields a denser, better-scaled gradient at the aggregation layer.  We provide details of our methodologies in the subsequent sections.

\subsection{Overview}

\begin{figure}[h]
\begin{center}
\centerline{\includegraphics[width=0.7\columnwidth]{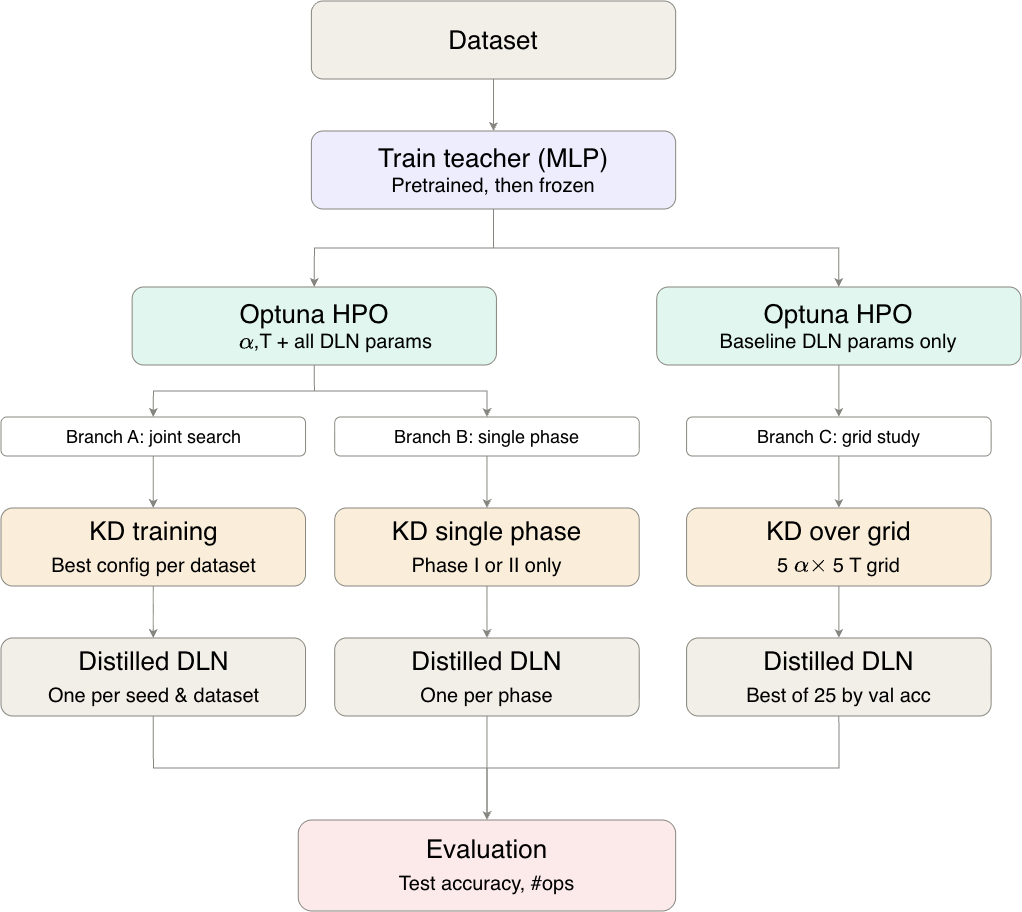}}
\caption{KD flowchart for DLNs.}
\label{methodology}
\end{center}
\end{figure}

We present a high-level overview of the distillation workflow in Fig.~\ref{methodology} and a general algorithmic summary for KD in Algorithm \ref{alg:kd-skeleton}. 

\begin{algorithm}[h]
\caption{Skeleton training process for DLN KD}
\label{alg:kd-skeleton}
\begin{algorithmic}[1]
\REQUIRE Number of iterations $T_{\text{iter}}$, epochs per phase $E$
\REQUIRE Training dataset $\mathcal{D}$, distillation temperature $T$, distillation weight $\alpha$
\STATE Train teacher network $f_{\text{T}}$ on $\mathcal{D}$ and freeze its parameters.
\FOR{$t = 1$ to $T_{\text{iter}}$}
    \STATE \{Phase I: Optimizing neuron functions\}
    \STATE Quantize and fix neuron connections.
    \FOR{$\text{epoch} = 1$ to $E$}
        \STATE Compute student logits $\mathbf{z}^{\text{s}}$ and teacher logits $\mathbf{z}^{\text{t}} = f_{\text{T}}(\mathbf{x})$.
        \STATE Compute loss $\mathcal{L}_{\text{KD}}$ from $\mathbf{z}^{\text{s}}$, $\mathbf{z}^{\text{t}}$, and the true label, using $\alpha$ and $T$.
        \STATE Apply gradient of $\mathcal{L}_{\text{KD}}$ to logic gate distributions for neurons
               in LogicLayer and to biases and slopes for neurons in
               ThresholdLayer. \{Update neuron parameters\}
    \ENDFOR
    \STATE \{Phase II: Optimizing neuron connections\}
    \STATE Quantize and fix functions inside neurons.
    \FOR{$\text{epoch} = 1$ to $E$}
        \STATE Compute student logits $\mathbf{z}^{\text{s}}$ and teacher logits $\mathbf{z}^{\text{t}} = f_{\text{T}}(\mathbf{x})$.
        \STATE Compute $\mathcal{L}_{\text{KD}}$ from $\mathbf{z}^{\text{s}}$, $\mathbf{z}^{\text{t}}$, and the true label, using $\alpha$ and $T$.
        \STATE Apply gradient of $\mathcal{L}_{\text{KD}}$ to find optimal connections between
               neurons. \{Update connection parameters\}
    \ENDFOR
\ENDFOR
\end{algorithmic}
\end{algorithm}

We first obtain a baseline accuracy for a regular DLN.  We conduct a hyperparameter search over the DLN architecture and training parameters, following the procedure of \citep{dln}, and record the best validation accuracy obtained. We then train an MLP teacher, scaling its capacity until its validation accuracy exceeds that of the tuned DLN. This ordering ensures that the teacher provides a target the student has not already reached. Then, we start the KD process by pretraining and freezing the MLP teacher for the dataset, as shown in Fig.~\ref{methodology}.  With the teacher fixed, we train DLN students using the distillation objective in place of cross-entropy, retaining the two-phase alternating schedule. We consider three search regimes for the distillation hyperparameters. For the first two, distillation weight $\alpha$ and temperature $T$ are included in the hyperparameter search alongside the DLN architecture parameters, so that the search selects them jointly with the model.  This is further split into Branch A, where the distillation is done over both phases, and Branch B, where distillation only occurs in a single phase (Phase I only or Phase II only).  In the third regime (Branch C), we fix the architecture at the configuration selected by the baseline search and sweep a grid of 25 ($\alpha$, $T$) combinations, isolating the effect of the two distillation hyperparameters. The former reflects how the method performs under realistic joint optimization, whereas the latter characterizes its sensitivity to $\alpha$ and $T$ independently of architectural confounds. We later evaluate the test accuracy and number of basic logic gate operations, and analyze which variant produces the best distilled DLN for each dataset.

Algorithm \ref{alg:kd-skeleton} provides a deeper look into the KD training process for distillation over both phases.  Each outer iteration alternates between Phase I that optimizes neuron operators (with frozen and hard-assigned connections) and Phase II that optimizes connections (with frozen and hard-assigned operators). During both phases, we replace the standard cross-entropy objective with a distillation objective that combines a hard-label cross-entropy term against the ground truth with a temperature-scaled soft-target term against the teacher, as shown in lines 6-8 and 13-15. For distillation over single phases, either one of these sections can be kept as the regular standard cross-entropy objective.  Because the distillation objective is evaluated at the network output, its gradient flows back through whichever parameters are active in the current phase; hence, distillation applies uniformly across both phases without any change to the phase schedule, the discretization of operators and connections, or the inference procedure. The teacher is a fixed model trained beforehand and scaled until it reaches a higher accuracy than a regular DLN.  It participates only in the forward pass during student training and is discarded at inference. As a result, the distilled DLN incurs no additional inference cost, parameters, or storage relative to a conventionally trained DLN.  We verify this property empirically alongside accuracy.

\subsection{Distillation Objective}

Training is differentiable through three relaxations. Each LogicLayer neuron $i$ holds operator logits $W_i \in \mathbb{R}^{16}$ and two link-logit vectors $U_i, V_i \in \mathbb{R}^{\text{in\_dim}}$, and their Softmaxes give a soft operator mixture and soft input selections, respectively. Each SumLayer entry holds a connection logit $S_{j,i}$, with $\sigma(S_{j,i})$ depicting the soft probability that last-layer neuron $j$ feeds class $i$. Training proceeds in $T_{\text{iter}}$ outer iterations, each alternating Phase I (optimize operator and threshold parameters with links frozen and hard-assigned) and Phase II (optimize link and SumLayer parameters with operators frozen and hard-assigned).

For an input, let the teacher emit logits $z^{t} \in \mathbb{R}^{C}$ and the DLN emit raw class scores $z^{s} \in \mathbb{R}^{C}$ from the SumLayer for a $C$-class problem. Because $z^{s}$ scores are integer-like rule counts whose range depends on the number of neurons feeding each class, we map them to logits with the DLN's count-normalization temperature $\tau$:

\begin{equation}
    \tilde{z}^{s}_i \;=\; \frac{z^{s}_i}{\tau}.
    \label{eq:count-norm}
\end{equation}

We use $\tau$ as a DLN hyperparameter and introduce a separate distillation temperature $T$. We treat the two as distinct hyperparameters throughout.

With distillation temperature $T$, we define the softened distributions as follows.
\begin{equation}
p^{t} = \mathrm{softmax}\big(z^{t}/T\big), \qquad p^{s} = \mathrm{softmax}\big(\tilde{z}^{s}/T\big),
\end{equation}
and train the student to minimize

\begin{equation}
\mathcal{L}_{KD} \;=\; \alpha T^{2} \mathrm{KL}\big(p^{t}\|p^{s}\big) \;+\; (1-\alpha) \mathrm{CE}\big(\mathrm{softmax}(\tilde{z}^{s}),y\big),
\end{equation}
where $y$ is the ground-truth label, $\alpha \in [0,1]$ is the distillation weight that balances the soft teacher signal against the hard label, and the $T^{2}$ factor restores the soft-target gradient to the same scale as the hard-target gradient. The teacher is trained once, prior to student training, and held fixed. Its parameters receive no gradient. It is evaluated with \text{torch.no\_grad} during the student's forward pass.

\begin{algorithm}[!ht]
\caption{Training with output-layer KD in both phases}
\label{alg:kd-both}
\begin{algorithmic}[1]
\REQUIRE Number of iterations $T_{\text{iter}}$, epochs per phase $E$
\REQUIRE Pretrained teacher $f_{\text{T}}$, distillation temperature $T$, distillation weight $\alpha$
\STATE Train teacher $f_{\text{T}}$ and freeze its parameters
\FOR{$t = 1$ to $T_{\text{iter}}$}
    \STATE \{Phase I: optimizing neuron functions\}
    \STATE Freeze link parameters $\mathbf{U}, \mathbf{V}, \mathbf{S}$.
    \FOR{$\text{epoch} = 1$ to $E$}
        \FORALL{$(\mathbf{x}, y)$ in $\mathcal{D}$}
            \STATE $\mathbf{\tilde{z}}^{\text{s}} \leftarrow$ forward pass in Phase I mode
            \STATE $\mathbf{z}^{\text{t}} = f_{\text{T}}(\mathbf{x})$ \{No gradient\}
            \STATE $\mathcal{L}_{KD} = \alpha T^2 \text{KL}\!\left(\text{Softmax}(\tfrac{\mathbf{z}^{\text{t}}}{T}) \big\| \text{Softmax}(\tfrac{\mathbf{\tilde{z}}^{\text{s}}}{T})\right)
                    + (1-\alpha)\mathcal{L}_{\text{CE}}(\text{Softmax}(\mathbf{\tilde{z}}^{\text{s}}), y)$
            \STATE Apply gradient of $\mathcal{L}_{KD}$ to $\mathbf{b}, \mathbf{s},\mathbf{W}$. \{Update neuron parameters\}
        \ENDFOR
    \ENDFOR
    \STATE \{Phase II: optimizing neuron connections\}
    \STATE Freeze neuron function parameters $\mathbf{b}, \mathbf{s}, \mathbf{W}$.
    \FOR{$\text{epoch} = 1$ to $E$}
        \FORALL{$(\mathbf{x}, y)$ in $\mathcal{D}$}
            \STATE $\mathbf{\tilde{z}}^{\text{s}} \leftarrow$ forward pass in Phase II mode
            % \IF{$\mathbf{\tilde{z}}^{\text{s}}$ requires no gradient}
            %     \STATE \textbf{continue} \{No update possible; skip teacher forward\}
            % \ENDIF
            \STATE $\mathbf{z}^{\text{t}} = f_{\text{T}}(\mathbf{x})$ \{No gradient\}
            \STATE $\mathcal{L}_{KD} = \alpha T^2 \text{KL}\!\left(\text{Softmax}(\tfrac{\mathbf{z}^{\text{t}}}{T}) \big\| \text{Softmax}(\tfrac{\mathbf{z}^{\text{s}}}{T})\right)
                    + (1-\alpha)\mathcal{L}_{\text{CE}}(\text{Softmax}(\mathbf{\tilde{z}}^{\text{s}}), y)$
            \STATE Apply gradient of $\mathcal{L}_{KD}$ to $\mathbf{U}, \mathbf{V}, \mathbf{S}$. \{Update connection parameters\}
        \ENDFOR
    \ENDFOR
\ENDFOR
\end{algorithmic}
\end{algorithm}

\subsection{Grid Search over Distillation Hyperparameters}
To characterize how the distillation weight $\alpha$ and temperature $T$ affect the student in isolation, we hold the DLN architecture fixed and vary only these two hyperparameters. We take the architecture from the configuration selected by baseline hyperparameter search of the original DLN, so that every point in the grid trains the same network and any difference in outcome is attributable to distillation rather than to a change in model capacity. We sweep a 5 $\times$ 5 grid of ($\alpha$, $T$) pairs, training one student per pair with the procedure shown in Algorithm \ref{alg:kd-both}, and record its validation accuracy. Because the architecture is fixed rather than re-searched and because this study is intended to expose the shape of the response surface rather than produce deployment models, we train each grid point under a single seed. The resulting accuracy surface indicates the region of ($\alpha$, $T$) in which distillation is beneficial and reveals whether a favorable temperature depends on the chosen $\alpha$, $T$ or whether the two act independently.  We choose the combination that produces the highest validation accuracy and use it on the test dataset to retrieve the final test accuracy for comparison with the other methods.

\subsection{Joint Hyperparameter Search with Distillation}
In this regime, we add $\alpha$ and $T$ to the search space alongside the DLN architecture and training parameters, and let the search optimize them jointly. Each trial samples a complete configuration with architecture and distillation hyperparameters together and trains a student under that configuration using the distillation procedure of Algorithm \ref{alg:kd-both}, in which the full distillation loss supervises both Phase I and Phase II. We score the trial by cross-validated balanced-class accuracy, as in the baseline search. Because the architecture is no longer held constant, the search may select a model that differs from the baseline configuration. Thus, we choose the distillation hyperparameters in the context of that model rather than in isolation. This regime measures the accuracy attainable when distillation is treated as an integral part of model selection. This is the setting under which we report the method's headline results. We repeat the search across four seeds and select the best outcome.

\subsection{Distillation in a Single Phase}
We compute the distillation loss at the network output. Its gradient reaches whichever parameters are unfrozen in the current phase. We explore whether the benefit of distillation arises from shaping the neuron functions, shaping the connections, or both (observed from the previous variant). To enable this, we consider two restricted variants of the full procedure. In the neuron-only variant, we use the distillation loss during Phase I, whereas Phase II reverts to the conventional cross-entropy objective, implying that the connections are optimized without a teacher.   Similarly, in the connection-only variant, distillation supervises Phase II, whereas Phase I is trained with cross-entropy alone. Both variants leave the two-phase schedule, the architecture, and the inference procedure unchanged, differing from the full method only in the phase that receives the teacher's signal. Comparing them against the full procedure and against the undistilled baseline isolates the contribution of each phase and indicates where distillation is most effective in the network.

%%%%%%%%%%%%% RESULTS %%%%%%%%%%%%%%%%
\section{Experimental Results}
We evaluate the proposed DLN KD method on 20 public datasets. 
%the same as those used in \citep{dln}. 
The baselines are DLNs without KD (best over four seeds). We assess predictive accuracy and inference cost (e.g., number of operations), and compare joint hyperparameter optimization (HPO), grid search, and each single-phase distillation method used to search for the best configuration and distilled model. In addition, we explain how KD can be used to decrease DLN training time and present a personalized interpretation using DLNs.

\begin{table*}[h]
\caption{Balanced-class test accuracies for DLNs obtained using KD.}
\label{accuracy-table}
\vskip 0.15in
\begin{center}
\begin{small}
\begin{sc}
\begin{tabular*}{\textwidth}{@{\extracolsep{\fill}}lcc|cccc@{}}
\toprule
& \multicolumn{2}{c|}{Original} & \multicolumn{4}{c}{Knowledge Distillation} \\
\cmidrule(lr){2-3} \cmidrule(lr){4-7}
Data set & MLP (teacher) & DLN & Joint HPO & Grid Search & Neuron Phase & Connection Phase \\
\midrule
BankChurn        & 85.20 & 78.64 & 78.13 & 77.48 & 75.91 & 79.03 \\
BreastCancer     & 77.80 & 65.55 & 67.37 & 67.37 & 62.89 & 63.17 \\
Cardio           & 99.20 & 97.72 & 99.07 & 97.88 & 97.30 & 97.80 \\
CKD              & 99.00 & 100.0 & 98.65 & 98.39 & 98.33 & 100.0 \\
Cirrhosis        & 58.80 & 45.62 & 47.20 & 51.49 & 47.97 & 53.99 \\
Compas           & 68.90 & 66.28 & 65.95 & 68.38 & 66.84 & 66.41 \\
CreditFraud      & 95.90 & 94.72 & 94.92 & 92.68 & 94.92 & 93.90 \\
Diabetes         & 94.90 & 84.30 & 86.10 & 82.13 & 85.82 & 78.84 \\
EyeMovements     & 56.70 & 55.18 & 56.21 & 58.70 & 56.71 & 56.28 \\
HAR              & 92.70 & 86.96 & 87.06 & 88.45 & 88.61 & 87.82 \\
Heart            & 83.80 & 79.71 & 79.71 & 82.21 & 75.66 & 86.62 \\
HeartFailure     & 73.30 & 67.03 & 59.93 & 72.55 & 66.67 & 62.01 \\
HepatitisC       & 98.10 & 91.36 & 95.89 & 85.73 & 93.63 & 88.03 \\
Letter           & 97.00 & 72.63 & 71.32 & 70.01 & 75.46 & 74.48 \\
Liver            & 79.30 & 59.38 & 64.02 & 74.93 & 59.92 & 61.21 \\
Mushroom         & 100.0 & 100.0 & 100.0 & 99.84 & 100.0 & 100.0 \\
PDAC             & 83.70 & 56.35 & 55.61 & 56.76 & 50.90 & 50.19 \\
SatIm            & 91.70 & 85.76 & 84.87 & 86.46 & 84.92 & 86.94 \\
Stroke           & 94.90 & 77.28 & 73.73 & 72.00 & 71.21 & 75.42 \\
TelcoChurn       & 78.90 & 74.19 & 76.88 & 74.06 & 73.90 & 76.75 \\
\midrule
Average Rank     & - & - & 2.450	& 2.375	& 2.825	& 2.350 \\
\bottomrule
\end{tabular*}
\end{sc}
\end{small}
\end{center}
\end{table*}

\begin{table*}[h]
\caption{Number of basic logic gate-level operations for inference, assuming float16 for floating-point and int16 for integer.}
\label{ops-table}
\vskip 0.15in
\begin{center}
\begin{small}
\begin{sc}
\begin{tabular*}{\textwidth}{@{\extracolsep{\fill}}lcc|cccc@{}}
\toprule
& \multicolumn{2}{c|}{Original} & \multicolumn{4}{c}{Knowledge Distillation} \\
\cmidrule(lr){2-3} \cmidrule(lr){4-7}
Data set & MLP (teacher) & DLN & Joint HPO & Grid Search & Neuron Phase & Connection Phase \\
\midrule
BankChurn        & 7.67M & 1.6K    & 4.6K   & 3.7K   & 2.6K  & 2.8K  \\
BreastCancer     & 93.7M & 358     & 1.3K   & 358    & 211   & 1.8K  \\
Cardio           & 13.1M & 109.2K  & 6.4K   & 6.5K   & 6.0K  & 4.9K  \\
CKD              & 29.7M & 5.1K    & 3.7K   & 1.6K   & 2.6K  & 2.7K  \\
Cirrhosis        & 9.10M & 2.8K    & 2.1K   & 5.1K   & 3.2K  & 1.3K  \\
Compas           & 7.43M & 1.9K    & 2.8K   & 3.7K   & 3.2K  & 3.4K  \\
CreditFraud      & 15.0M & 8.1K    & 11.2K  & 23.9K  & 21.0K & 6.4K  \\
Diabetes         & 9.66M & 5.0K    & 3.6K   & 2.0K   & 959   & 1.6K  \\
EyeMovements     & 11.7M & 11.5K   & 182.2K & 16.5K  & 14.8K & 10.8K \\
HAR              & 25.8M & 46.5K   & 30.6K  & 52.6K  & 48.7K & 58.7K \\
Heart            & 8.14M & 1.6K    & 1.7K   & 825    & 1.0K  & 2.4K  \\
HeartFailure     & 7.75M & 3.0K    & 2.0K   & 3.2K   & 2.1K  & 1.4K  \\
HepatitisC       & 9.02M & 6.9K    & 4.7K   & 6.0K   & 2.3K  & 2.3K  \\
Letter           & 12.4M & 23.5K   & 23.0K  & 20.5K  & 28.3K & 30.0K \\
Liver            & 8.38M & 3.7K    & 5.1K   & 4.6K   & 2.8K  & 1.3K  \\
Mushroom         & 10.8M & 2.6K    & 3.1K   & 1.9K   & 2.5K  & 2.5K  \\
PDAC             & 15.3M & 309.7K  & 2.5K   & 117.3K & 2.1K  & 2.9K  \\
SatIm            & 17.2M & 32.4K   & 24.6K  & 18.9K  & 32.8K & 22.0K \\
Stroke           & 7.27M & 1.3K    & 2.8K   & 1.2K   & 469   & 1.9K  \\
TelcoChurn       & 7.75M & 15.3K   & 21.5K  & 1.7K   & 976   & 2.2K  \\
\midrule
Average Rank     & - & - & 2.95	& 2.70 & 2.00 & 2.35 \\
\bottomrule
\end{tabular*}
\end{sc}
\end{small}
\end{center}
\end{table*}

\begin{table*}[h]
\caption{Number of high-level operations for inference, assuming float16 for floating-point and int16 for integer.}
\label{high-ops-table}
\vskip 0.15in
\begin{center}
\begin{small}
\begin{sc}
\begin{tabular*}{\textwidth}{@{\extracolsep{\fill}}lcc|cccc@{}}
\toprule
& \multicolumn{2}{c|}{Original} & \multicolumn{4}{c}{Knowledge Distillation} \\
\cmidrule(lr){2-3} \cmidrule(lr){4-7}
Data set & MLP (teacher) & DLN & Joint HPO & Grid Search & Neuron Phase & Connection Phase \\
\midrule
BankChurn        & 12.4K  & 44   & 162  & 99    & 58  & 70  \\
BreastCancer     & 151.0K & 82   & 511  & 83    & 58  & 683 \\
Cardio           & 21.1K  & 1839 & 179  & 250   & 296 & 206 \\
CKD              & 47.6K  & 283  & 91   & 105   & 62  & 73  \\
Cirrhosis        & 14.7K  & 38   & 37   & 135   & 123 & 23  \\
Compas           & 12.0K  & 39   & 54   & 71    & 100 & 92  \\
CreditFraud      & 24.2K  & 87   & 129  & 865   & 367 & 71  \\
Diabetes         & 15.6K  & 83   & 42   & 26    & 9   & 25  \\
EyeMovements     & 18.9K  & 392  & 1908 & 540   & 291 & 223 \\
HAR              & 41.6K  & 573  & 337  & 1176  & 642 & 862 \\
Heart            & 13.2K  & 78   & 32   & 76    & 30  & 210 \\
HeartFailure     & 12.5K  & 44   & 31   & 58    & 51  & 14  \\
HepatitisC       & 14.6K  & 161  & 61   & 123   & 22  & 26  \\
Letter           & 20.1K  & 220  & 170  & 222   & 252 & 246 \\
Liver            & 13.6K  & 58   & 86   & 66    & 33  & 14  \\
Mushroom         & 17.4K  & 865  & 1124 & 563   & 874 & 874 \\
PDAC             & 24.7K  & 3043 & 28   & 1818  & 23  & 34  \\
SatIm            & 27.8K  & 567  & 489  & 343   & 541 & 303 \\
Stroke           & 11.8K  & 36   & 71   & 60    & 14  & 41  \\
TelcoChurn       & 12.5K  & 865  & 775  & 108   & 78  & 134 \\
\midrule
Average Rank     & - & - & 2.70 & 3.00 & 2.125 & 2.175 \\
\bottomrule
\end{tabular*}
\end{sc}
\end{small}
\end{center}
\end{table*}

\subsection{Experimental Setup}

We process the datasets by removing samples with many missing values, one-hot encoding categorical features, and scaling continuous features with Min-Max scalers. For the baseline, joint HPO, and single-phase distillation searches, we run four independent trials, each with a different random seed. For grid search, we run an independent trial with a random seed for each dataset-model pair.  Each trial has three stages: hyperparameter search, training, and evaluation. For hyperparameter search, we employ Optuna \citep{akiba2019optuna} to sample 32 hyperparameter sets from a predefined grid and select the value configuration that yields the best cross-validated balanced-class test accuracy. Besides the grid search, the other KD hyperparameter searches include the $\alpha$ and $T$ hyperparameters.  Depending on the dataset size, we use two to four folds, with smaller datasets receiving more folds. Data processing relies on scikit-learn \citep{pedregosa2011scikit}, whereas the MLP and DLN models are implemented in PyTorch \citep{pytorch}. The teacher MLPs employ a 3-layer MLP with 1D batch-norm, with a hidden layer size of 128.  This hidden layer size enables most MLPs to reach accuracies surpassing that of DLNs.  Ray \citep{moritz2018ray} orchestrates parallel hyperparameter searches and training runs.

\subsection{Accuracy}
We measure predictive performance using balanced-class accuracy. Table \ref{accuracy-table} presents the results of the baseline DLN, the four distillation variants (grid search, joint HPO for both-phase distillation, neuron-only distillation, and connection-only distillation), and the MLP teacher across the 20 datasets. No single distillation variant dominates: The best-performing method varies from dataset to dataset; the improvement it yields over the baseline DLN ranges from negligible to substantial.  Overall, the average rank indicates that the connection-only distillation yields the highest accuracies among the four variants followed by grid search, joint, and neuron-only distillation methods.

On only the Stroke dataset, distillation offers no benefit. None of the three variants improves on the DLN baseline, indicating that a stronger teacher signal does not always translate into a better student. Where distillation does help, the magnitude of the gain varies widely. For the PDAC dataset, the best variant (grid search) improves accuracy by only 0.41\%, whereas for the Liver dataset, the grid search method finds a model that raises the accuracy by 15.55\%, which is the largest improvement we observe. Between these extremes lies a range of gains that depend on both the dataset and the variant.

We further observe that on the EyeMovements and Heart datasets, the distilled DLN surpasses not only the baseline but also the MLP teacher from which it is distilled. This indicates that distillation is not acting purely as a channel for transferring the teacher's decision function. %On these datasets the student attains an accuracy the teacher itself did not reach, 
This suggests that the softened teacher targets serve as a form of regularization or improved training signal rather than an upper bound the student can approach.

\subsection{Efficiency}

\begin{figure}[h]
\begin{center}
\centerline{\includegraphics[width=\columnwidth]{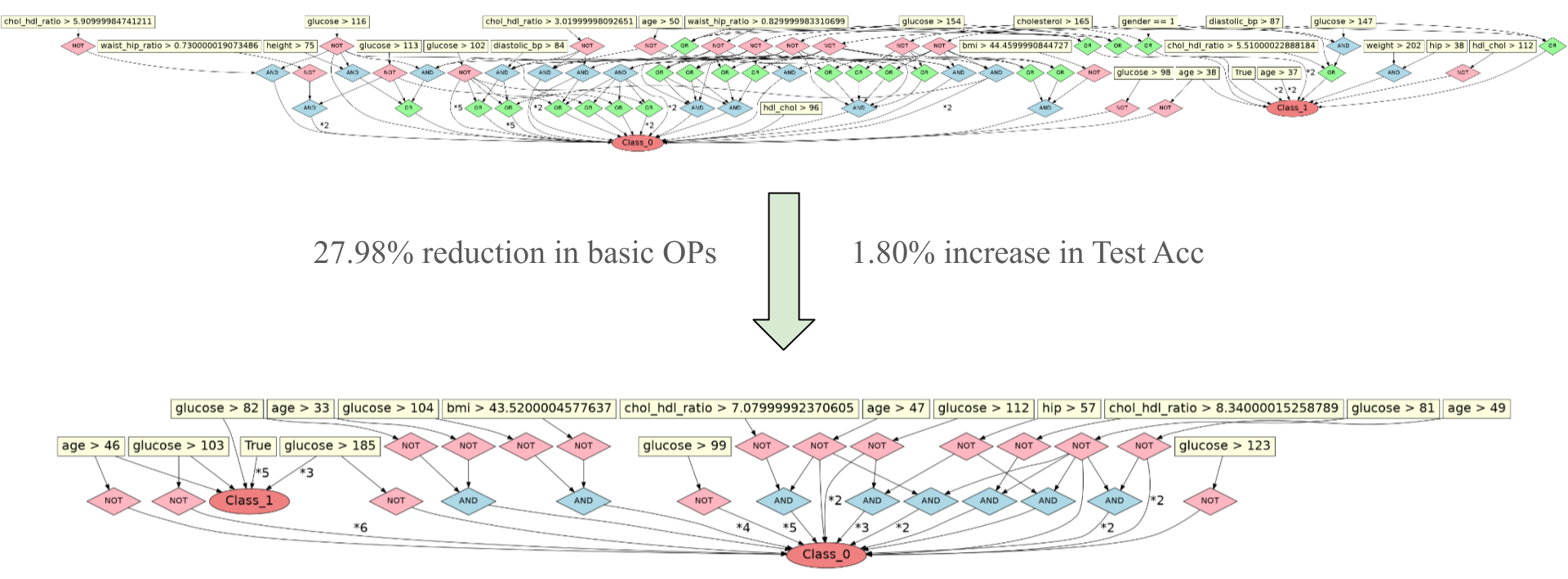}}
\caption{The difference in model size for the Diabetes dataset can be observed. The top model is the original DLN model, and the bottom model is the distilled DLN model.}
\label{reduction_ex}
\end{center}
\end{figure}

Next, we measure inference cost as the number of basic hardware logic-gate operations (OPs). We decompose high-level computations, such as floating-point additions and multiplications, into fundamental logic gates. In this metric, a two-input AND, OR, NAND, or NOR counts as one OP, and a two-input XOR or XNOR counts as three OPs, whereas a NOT gate counts as 0 OPs. While training utilizes PyTorch and scikit-learn defaults (float32 or float64), our cost analysis assumes float16 for floating-point and int16 for integer arithmetic, since nearly all models maintain accuracy at these precisions.

Table \ref{ops-table} reports the gate-level operation counts and Table \ref{high-ops-table} the number of high-level operations for the baseline DLN, the four distillation methods, and the MLP teacher. The central observation is that distillation does not materially change the size of the resulting model. For most datasets, the distilled DLN requires a number of basic logic gates and high-level operations comparable to that of a conventionally trained DLN. As with accuracy, the most efficient distillation method depends on the dataset; no single variant yields the smallest DLNs across the board.

The exceptions fall in both directions. In some cases, the distilled model is larger. For example, the gate count of the best model for the Compas dataset increases by about 95\% relative to the baseline DLN. Even then, however, it remains three orders of magnitude below the gate count of the corresponding MLP, so that the efficiency advantage of the logic network over a conventional neural network is preserved. In the opposite direction, distillation can yield a model that is both more accurate and smaller. On the Diabetes dataset, distillation (joint HPO) reduces the gate count to about half that of the baseline DLN while improving accuracy by 1.80\%.  Fig. \ref{reduction_ex} shows the noticeable decrease in model size for the Diabetes dataset.  The top model is the baseline DLN, and the bottom model is the distilled DLN. We also observe, for the CKD and Mushroom datasets, that although the best model accuracy matches that of the baseline DLN, the distilled DLNs have a smaller number of operations and size.

The neuron-only and connection-only variants rank better on efficiency, whereas the joint HPO and grid search methods rank worse. Across all datasets, the gate counts of all four methods remain within the same order of magnitude as the baseline DLN and far below that of the MLP. Hence, the accuracy gains come at no meaningful expense to efficiency.

\subsection{Grid Search Response Surface}

\begin{figure}[h]
\begin{center}
\centerline{\includegraphics[width=0.6\columnwidth]{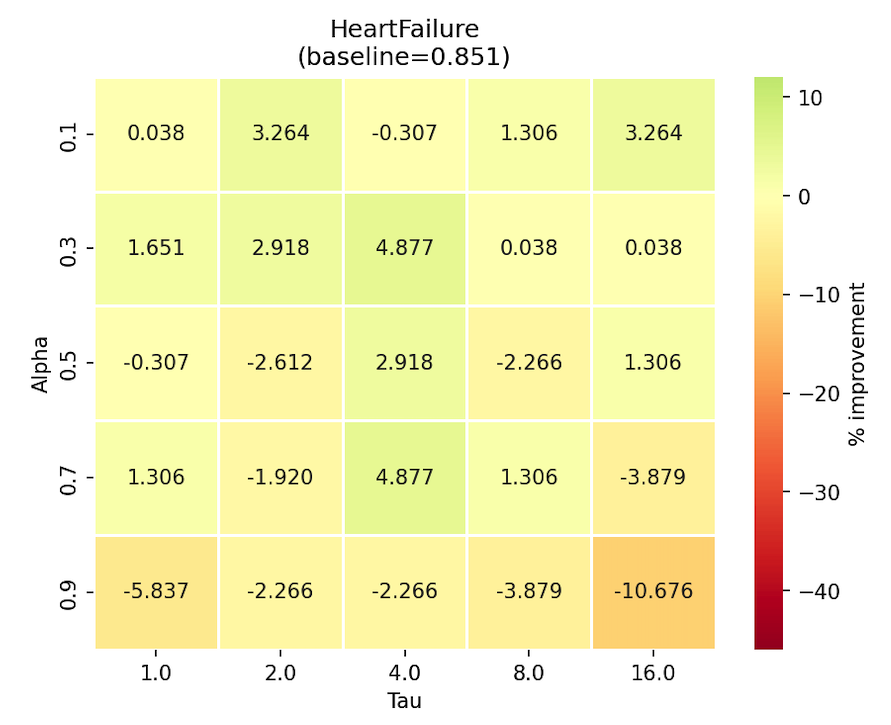}}
\caption{An example heatmap for the HeartFailure dataset for grid search.  The percentage improvement in validation accuracy compared to the baseline DLN validation accuracy is recorded over 25 combinations of $\alpha$ and $T$.}
\label{grid_search}
\end{center}
\end{figure}

For each dataset, we record the validation accuracy for each $\alpha$, $T$ pair and compare it with the baseline DLN validation accuracy.  A heatmap, such as the one shown in Fig.~\ref{grid_search}, illustrates how various combinations perform. We find no correlation or pattern across the surfaces for all the datasets, implying that the best combination is unique for each dataset, and that $\alpha$ and $T$ work independently.  We also note that not all combinations yield increases in accuracy; hence, exploration of this space is important to obtain a well-distilled model.

\subsection{Decrease in Training Time with KD}

Because KD can be used to increase accuracy, it can also be used to decrease the overall DLN training time for large datasets.  The most time-consuming part of DLN training is the hyperparameter search, which is crucial to finding a higher accuracy model.  However, if we use KD, it may be possible to skip hyperparameter search and still obtain a model with comparable accuracy.  

\begin{figure}[h]
\begin{center}
\centerline{\includegraphics[width=1.0\columnwidth]{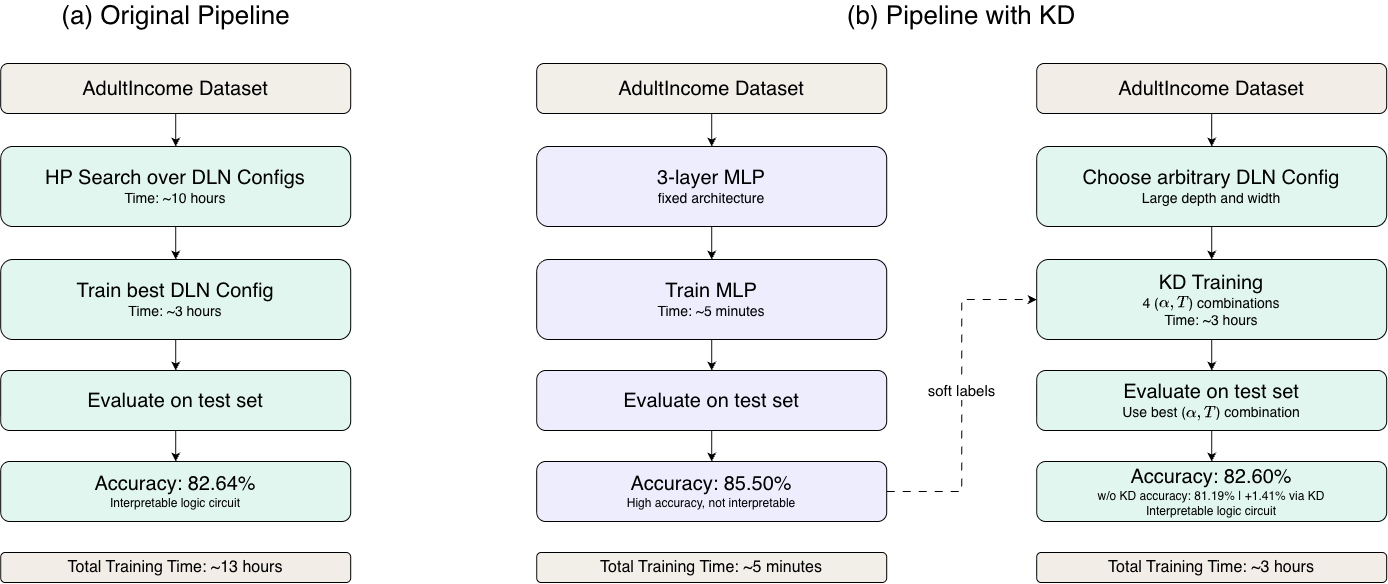}}
\caption{A comparison of training pipelines and time using the AdultIncome dataset.  (a) This is the original DLN training pipeline, which takes about 13 hours to achieve 82.64\% accuracy. (b) The KD pipeline that does not employ  hyperparameter search takes 3 hours to achieve a similar accuracy to (a).}
\label{adult_pipeline}
\end{center}
\end{figure}

We provide an example, a proof-of-concept, in Fig. \ref{adult_pipeline}.  This example uses the UCI AdultIncome dataset, which is much larger than most of the 20 datasets used before.  Fig.~\ref{adult_pipeline}(a) shows a typical training pipeline. Because of the size of the dataset, hyperparameter search and training takes about a total of 13 hours  to find a DLN model with an accuracy of 82.64\% for one seed.  Fig.~\ref{adult_pipeline}(b) shows the KD pipeline that does not employ hyperparameter search.  Instead, it chooses an arbitrary DLN configuration with relatively large depth and width.  Because we use SymPy \citep{meurer2017sympy} at the end to simplify all the logic expressions, the size during training does not significantly affect the size of the final DLN model.  Separately, a teacher MLP is also trained on the dataset, which takes about 5 minutes and reaches a test accuracy of 85.50\%.  The soft labels from this model are then used for KD.  Because we do not know the best ($\alpha, T$) combination for distillation, we try four different combinations.  They can be run in parallel and take about 3 hours to train.  The combination that distills a DLN model with the best validation accuracy is used for evaluation on the test set.  The total training time becomes about 3 hours and the final distilled model has a test accuracy of 82.60\%, which is only 0.04\% less than the DLN model found in the typical training pipeline in part (a).  Thus, it is possible to use KD to decrease the total training time for a DLN model, while also achieving comparable accuracy.

\subsection{Personalized Interpretation}

\begin{figure}[h]
\begin{center}
\centerline{\includegraphics[width=\columnwidth]{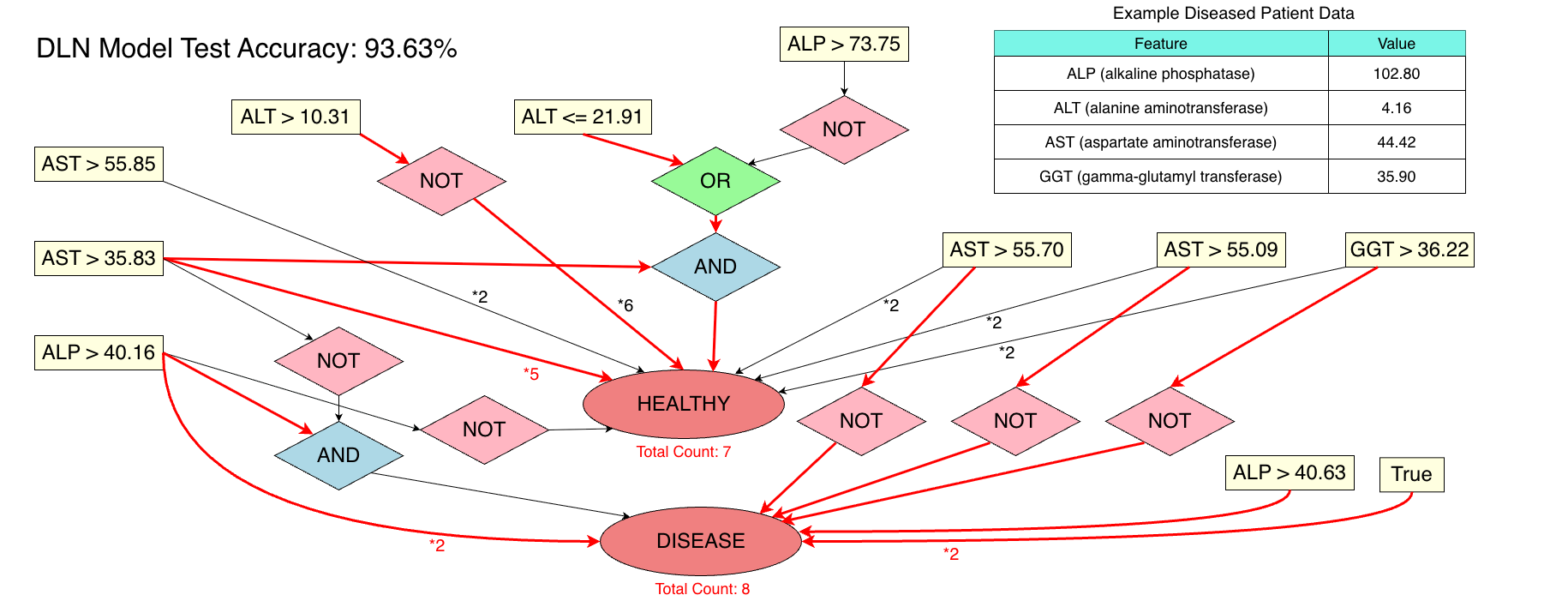}}
\caption{An example DLN model for the HepatitisC dataset.  Given some feature values for an example diseased patient, the triggered paths are indicated in red.  It is possible to provide a personalized interpretation to why this patient is labeled as diseased by tracing back the red lines to specific conditions (e.g., ALP $>$ 40.63).}
\label{interpretation_graph}
\end{center}
\end{figure}

Because both its gates and connections are trained, a DLN is very compact and relatively easy to trace.  We provide an example of how DLNs can be used for personalized interpretation of model results in Fig.~\ref{interpretation_graph}.  We show a DLN model trained on the HepatitisC dataset with a test accuracy of 93.63\%.  We use an instance of a diseased patient from the dataset as an illustrative example.  Although there are originally 12 features, because the model only needs four, we only include these four features.  The red lines indicate the paths that are triggered by this data instance.  After summing these paths, the 'healthy' result has a total count of 7 while the 'disease' result has a total count of 8.  Because the 'disease' sum is higher, the model concludes that this patient is diseased.

On top of the correct final classification, we show that it is possible to interpret why the model came to this conclusion for this specific patient.  Tracing the red lines back from `disease,' we can find the conditions that were triggered.  In Fig. \ref{interpretation_graph}, the triggered conditions are `ALP $>$ 40.16', `ALP $>$ 40.62', `AST $\leq$ 55.70', `AST $\leq$ 55.09', and `GGT $\leq$ 36.22'.  Altogether, we can say that the DLN model classifies the patient as diseased because the alkaline phosphatase level is over 40.62 U/L, aspartate aminotransferase level is less than 55.09 U/L, and  gamma-glutamyl transferase is less than 36.33 U/L.  Thus, DLNs make the result interpretable in addition to providing high classification accuracy for each patient.

%%%%%%%%%%%% DISCUSSIONS %%%%%%%%%%%%%%%%%
\section{Conclusion}
We investigated KD as a means for improving the accuracy of DLNs. Motivated by the vanishing-gradient behavior of these networks, we distilled a DLN student from a trained MLP teacher, integrating the distillation objective into the two-phase training procedure without altering the architecture or the inference cost of the model. We evaluated the method through joint HPO, grid search over the distillation hyperparameters, and two single-phase variants, across 20 tabular datasets.

Our results indicate that distillation has the potential to improve DLN accuracy. 
Among the variants we studied, treating the distillation hyperparameters as part of model search, whether through joint HPO or grid search, was the most effective route to higher accuracy.
The gains are obtained at negligible cost to gate-level efficiency. We also showed that KD can help decrease the total training time and provide personalized interpretability through DLNs.

Several research directions remain open. 
A distillation objective designed specifically for DLN structure instead of standard logit-based loss might yield more reliable gains. Thus, identifying such an objective is a natural next step. The choice of teacher is similarly unexplored. Teachers, whose decision structure is more compatible with a logic network, may transfer their knowledge more effectively.  Using a DLN teacher and exploring self-distillation in the born-again setting is another possible extension. In addition, matching not only the output of the teacher but also the outputs of its intermediate stages could enhance the distillation process.

More broadly, distillation of a neural network into a DLN is one instance of transferring the capability of scalable but opaque models into architectures that are interpretable by design. 

\vspace*{2mm}
\noindent
\textbf{Software Availability:} The code can be found at \href{https://github.com/hys-1364/kd-dln}{this link}.

\vspace*{2mm}
\noindent
\textbf{Acknowledgments:} This work was supported by NSF under Grant No. CCF-2416541.

\bibliographystyle{unsrtnat}
\bibliography{references}  %%% Uncomment this line and comment out the ``thebibliography'' section below to use the external .bib file (using bibtex) .





% \end{thebibliography}

\end{document}